\documentclass[preprint,authoryear,12pt]{elsarticle}
\usepackage{amsfonts}
\usepackage{amsmath}
\usepackage{graphicx}
\usepackage{natbib}
\usepackage{url}
\usepackage{float}
\usepackage{subfigure}
\usepackage[graphicx]{realboxes}
\usepackage{epstopdf}%
\usepackage[british]{babel}
\usepackage{csquotes}

\usepackage{lineno,hyperref}
\modulolinenumbers[5]

\journal{Expert Systems with Applications}

\begin{document}
\begin{frontmatter}

\title{Automated Discovery of Mathematical Definitions in Text with Deep Neural Networks}

\author{Natalia Vanetik, Marina Litvak, Sergey Shevchuk, and Lior Reznik}
\address{Software Engineering Department, Shamoon College of Engineering, Bialik 56, Beer Sheva, Israel}
\ead{natalyav@sce.ac.il, marinal@sce.ac.il, sergesh@sce.ac.il, liorrenz@gmail.com}

\begin{abstract}
Automatic definition extraction from texts is an important task that has numerous applications 
in several natural language processing fields such as summarization, analysis of scientific texts, automatic taxonomy generation, ontology generation, concept identification, and question answering.
For definitions that are contained within a single sentence, this problem can be viewed as a binary classification of sentences into definitions
and non-definitions. 
In this paper,
we focus on automatic detection of one-sentence definitions in mathematical texts, which are difficult to separate from surrounding text. 
We experiment with several data representations, which include sentence syntactic structure and word embeddings, and apply deep learning methods such as the  Convolutional Neural Network (CNN) and the Long Short-Term Memory network (LSTM),
in order to identify mathematical definitions. 
Our experiments demonstrate the superiority of CNN and its combination with LSTM, when applied on the syntactically-enriched input representation. 
%
We also present a new dataset for definition extraction from mathematical texts. 
We demonstrate that this dataset is beneficial for training supervised models aimed at extraction of mathematical definitions. 
Our experiments with different domains demonstrate that mathematical definitions require special treatment, and that using cross-domain learning is inefficient for that task.
\end{abstract}

\end{frontmatter}

\section{Introduction}\label{sec:introduction}
Definitions have a very important role in scientific literature because they define the major concepts with which an article operates.
They are used in many automatic text analysis tasks, such as question answering, ontology matching and construction, formal concept analysis, and text summarization.
Intuitively, definitions are basic building blocks of a scientific article that are used to help properly describe hypotheses, experiments, and analyses.
It is often difficult to determine where a certain definition lies in the text because other sentences around it may have similar style. 
Automatic definition extraction (DE) is an important field in natural language processing (NLP) because it can be used to improve text analysis and search.

Definitions play a key role in mathematics, but their creation and use differ from those of \enquote*{everyday language} 
definitions. A comprehensive study is given in a series of works by Edwards and Ward~\citep{edwards2008undergraduate}, \citep{ edwards2004surprises}, \citep{edwards1998undergraduate}, 
 inspired by writings of Richard Robinson~\citep{robinson1962} and lexicographer Sidney Landau~\citep{landau2001}. 


Mathematical definitions frequently have a history as they evolve over time.
The definition we use for \textit{function}, for instance, may not be the one that was used a hundred years ago. 
The concept of \textit{connectivity} has two definitions, one for \textit{path connectivity} and another for \textit{set-theoretic connectivity}. In mathematical texts the meaning of the defined concept is not determined by its context but it is declared and is expected to have no variance within that specific mathematical text~\citep{edwards2008undergraduate}. 

Mathematical definitions have many features, some critical and some optional but accepted within the  mathematical community. 
\cite{van2003many} describe a good mathematical definition as containing criteria of \textit{hierarchy}
, \textit{existence}
, \textit{equivalence}, and \textit{axiomatization}. Desired but not necessary criteria of a definition are \textit{minimality}, \textit{elegance}, and \textit{degenerations}. 
We give here short definitions of these concepts; detailed explanations with examples can be found in \cite{van2003many}.
\begin{itemize}
\item \textbf{Hierarchy}: any new concept
must be described as a special case of a more general concept.
\item \textbf{Equivalence}: when one gives more than one formulation for the same concept, one must prove that they are equivalent.
\item \textbf{Axiomatization}: the definition fits in and is part of a deductive system.
\item \textbf{Minimality}: no more properties of the concept are mentioned in the definition than is required for its
existence.
\item \textbf{Elegance}: an elegant definition looks nicer, needs fewer words or less symbols, or uses more general basic concepts from which
the newly defined concept is derived.
\item \textbf{Degeneration}: what occurs at instances when our intuitive idea of a concept does not conform to a definition. 
\end{itemize}
Not every definition appearing in text is mathematical in the above sense. For example, Wikipedia articles contain definitions of different style.
We see below that the Wikipedia definition of the Kane \& Abel musical group 
is not similar in style to the Wikipedia definition of an Abelian group.
$$\begin{array}{l}
\mbox{\rm Definition 1: }\\
\mbox{\it \textbf{Kane \& Abel}, formally known as 'Double Vision', is an }\\
\mbox{\it American hip hop duo formed by twin brothers Daniel and }\\
\mbox{\it David Garcia that were founded by Master P in late 1995. }\\
\mbox{\it They were best known for their time with No Limit Records.}\\\\
\mbox{\rm Definition 2: }\\
\mbox{\it In abstract algebra, an \textbf{abelian group}, also called a}\\
\mbox{\it \textbf{commutative group}, is a group in which the result of }\\
\mbox{\it applying the group  operation to two group elements does  }\\
\mbox{\it not depend on the order in which they are written. }
\end{array}$$
Naturally, we expect to find  mathematical definitions in 
mathematical articles.  Mathematical definitions usually use formulas and 
notations extensively, both in definitions and in surrounding text. The number of words in mathematical text is smaller than in regular text due to the formulas that are used to express the former, but a presence of formulas is not a good indicator of a definition sentence because the surrounding sentences also use notations and formulas.
As an example of such text, Definition 3, below, contains a definition from Wolfram MathWorld. 
Only the first sentence in this text is considered a definition sentence.
$$\begin{array}{l}
\mbox{\rm Definition 3: }\\
\mbox{\it  Also called Macaulay ring, a \textbf{Cohen Macaulay ring} is a }\\
\mbox{\it Noetherian commutative unit ring R in which any proper }\\
\mbox{\it ideal $I$ of height $n$ contains a sequence $x_1,\dots,x_n$ of } \\
\mbox{\it elements (called a ring regular sequence) such that for all }\\
\mbox{\it $i=1,\dots, n$, the residue class of $x_i$ in the quotient ring }\\
\mbox{\it $R/\langle x_{1},\dots,x_{i-1}\rangle$ is a non-zero divisor. If $x_1,\dots, x_n$}\\
\mbox{\it  are indeterminate over a field $K$, the above condition}\\
\mbox{\it  is fulfilled by the maximal ideal $I=<x_1,\dots,x_n>$.}
\end{array}$$

Current methods for automatic DE view it as a binary classification task, 
where a sentence is classified as a definition or a non-definition. A supervised learning process is usually employed for this task, 
employing feature engineering for sentence representation. 
The absolute majority of current methods study generic definitions and not mathematical definitions (see Section \ref{rel-sec}).

In this paper we describe a supervised learning method for automatic DE from mathematical texts. Our method  applies a Convolutional Neural Network (CNN), a Long Short-Term Memory network (LSTM), and their combinations to the raw text data and sentence syntax structure, in order to detect definitions.
Our method is evaluated on three different corpora; two are well-known corpora for generic DE and one is a new annotated corpus of mathematical definitions, introduced in this paper.   

The main contributions of this paper are (1) analysis and introduction of the new annotated dataset of mathematical definitions, (2) evaluation of the state-of-the-art DE approaches on the new mathematical dataset, (3) introduction and evaluation of upgraded sentence representations adapted to mathematical domain with an adaptation of deep neural networks to new sentence representations, (4) extensive experiments with multiple network and input configurations (including  different embedding models) performed on different datasets in mathematical and non-mathematical domains, (5) experiments with   cross-domain and multi-domain learning in a DE task, and (6) introduction of the new parsed but non-annotated dataset composed of Wiki articles on  near-mathematics topics, used in an additional--extrinsic--evaluation scenario. These all contribute to showing that using specifically suited training data along with adapting sentence representation and classification models to the task of mathematical DE significantly improves extraction of mathematical definitions from surrounding text.

The paper is organized as follows. Section \ref{rel-sec} contains a survey of up-to-date related work. Section \ref{met-sec} describes the sentence representations and the structure of neural networks used in our approach. Section \ref{ev-sec} provides the description of the datasets, evaluation results, and their analysis. Section \ref{conc-sec} contains our conclusions. Finally, Appendix contains some supplementary  materials -- annotation instructions, description of the Wikipedia experiment, and figures.
\section{Related Work\label{rel-sec}}
Definition extraction has been a popular topic in NLP research for  more than a decade~\cite{xu2003trec}, and it remains a challenging and popular task today as a recent research call at SemEval-2020  shows~\footnote{\url{https://competitions.codalab.org/competitions/20900}}. Prior work in the field of DE can be divided into three main categories: (1) rule-based methods, (2) machine-learning methods relying on manual feature engineering, and (3) methods that use deep learning techniques.

Early works about DE from text documents belong to the first category. These works rely mainly on manually crafted rules based on linguistic parameters.
\cite{klavans2001evaluation} presented the DEFINDER, a rule-based system that mines consumer-oriented full text articles in order to extract definitions and the terms they define; the system is
evaluated on definitions from on-line dictionaries such as the UMLS Metathesaurus \citep{schuyler1993umls}.
\cite{xu2003trec} used various linguistic tools to  extract kernel facts for the definitional question-answering task in TREC 2003. 
\cite{malaise2004detecting} 
utilized semantic relations in order to mine defining expressions in domain-specific
corpora, thus detecting  semantic relations between the main terms in definitions. This work is evaluated on corpora from fields of anthropology and dietetics.
\cite{saggion2004mining,saggion2004identifying} employed analysis of on-line sources in order to
find lists of relevant secondary terms that frequently occur together with the definiendum in definition-bearing passages.
\cite{storrer2006automated} proposed a system that automatically detects and annotates definitions for technical terms in German text corpora.
Their approach focuses on verbs that typically appear in definitions by specifying search patterns based on the valency frames of definitor verbs.
\cite{borg2009evolutionary} extracted definitions from nontechnical texts by using genetic programming to learn the typical linguistic forms of definitions and then using a genetic algorithm to learn the relative importance of these forms. Most of these methods suffer from both low recall and precision (below $70\%$), because definition  sentences 
occur in highly variable and noisy syntactic structures. 

The second category of DE algorithms relies on semi-supervised and supervised machine learning that use semantic and other features
to extract definitions. This approach generates DE rules automatically but relies on feature engineering to do so. 
\cite{fahmi2006learning} presented an approach to learning concept definitions from
fully parsed text with a maximum entropy classifier incorporating various syntactic features; they tested this approach on a subcorpus of the
Dutch version of Wikipedia.
In \cite{westerhout2007combining}, a pattern-based glossary candidate detector, which is capable of extracting
definitions in eight languages, was presented.
\cite{westerhout2009definition} described a combined approach that first filters corpus with a definition-oriented grammar, and then applies machine learning 
to improve the results obtained with the grammar. The proposed algorithm was evaluated on a collection
of Dutch texts about computing and e-learning.
\cite{navigli2010learning} used Word-Class Lattices (WCLs), a generalization of word lattices, to model textual definitions. Authors introduced a new dataset called WCL that was used for the experiments. They achieved a $75.23\%$ F1 score on this dataset. 
\cite{reiplinger2012extracting} compared lexico-syntactic pattern bootstrapping and deep analysis. The manual rating experiment suggested that the concept of definition quality in a specialized domain is largely subjective, with a $0.65$ agreement score between raters. 
The DefMiner system, proposed in \citep{jin2013mining}, used Conditional Random Fields (CRF) to predict the function of a word and to determine whether this word is a part of a definition. The system was evaluated on a W00 dataset~\citep{jin2013mining}, which is a manually annotated subset of ACL-ARC ontology.
\cite{boella2013extracting} proposed a technique that
only uses syntactic dependencies between
terms extracted with a syntactic parser and then transforms
syntactic contexts to abstract representations in order to use a Support
Vector Machine (SVM).
\cite{anke2015weakly} proposed a weakly
supervised bootstrapping approach for identifying textual definitions with higher linguistic variability.
\cite{espinosa2014applying} presented a supervised approach to DE in which only
syntactic features derived from dependency relations are used.

Algorithms in the third category use Deep Learning (DL) techniques for DE, often incorporating syntactic features into the network structure.
\cite{li2016definition} used Long Short-Term Memory (LSTM) and word vectors to identify definitions and then tested this approach on the English and Chinese texts. Their method achieved a $91.2\%$ F-measure on the WCL dataset.
\cite{anke2018syntactically} combined CNN and LSTM, based on syntactic features and word vector representation of sentences. Their experiments showed the best F1 score ($94.2\%$) on the WCL dataset for CNN and the best F1 score ($57.4\%$) on the W00 dataset for the CNN and bidirectional LSTM (BLSTM) combination, both with syntactically-enriched sentence representation. 
Word embedding, when used as the input representation, have been shown to boost the performance in many NLP tasks, due to its ability to encode semantics. We believe, that a  choice to use word vectors as input representation in many DE works was motivated by its success in NLP-related classification tasks. 

We use the approach of \citep{anke2018syntactically} as a starting point and as a baseline for our method. 
We further extend this work by additional syntactic knowledge in a sentence representation model 
and by testing additional network architectures on our input. Due to observed differences in grammatical structure between regular sentences and definition sentences, we hypothesize that dependency parsing can add valuable features to their representation. Because extending a representation model results in a larger   input, we also hypothesize that a standard convolutional layer can help to automatically extract the most significant features before performing the classification task. Word embedding matrices enhanced with dependency information naturally call for CNN due to their size and CNN's ability to decrease dimensionality swiftly. On other hand, sentences are sequences for which LSTM is very suitable. In order to test the architecture properly, we needed to check how the order of these layers affects the results, and also to make sure that both layers are necessary. In order to test our hypothesis, we tested two variants of combined networks---LSTM and CNN---in different configurations on our data.    

\section{Methodology}\label{met-sec}
\begin{figure}[!t]
\center
\includegraphics[scale=0.4]{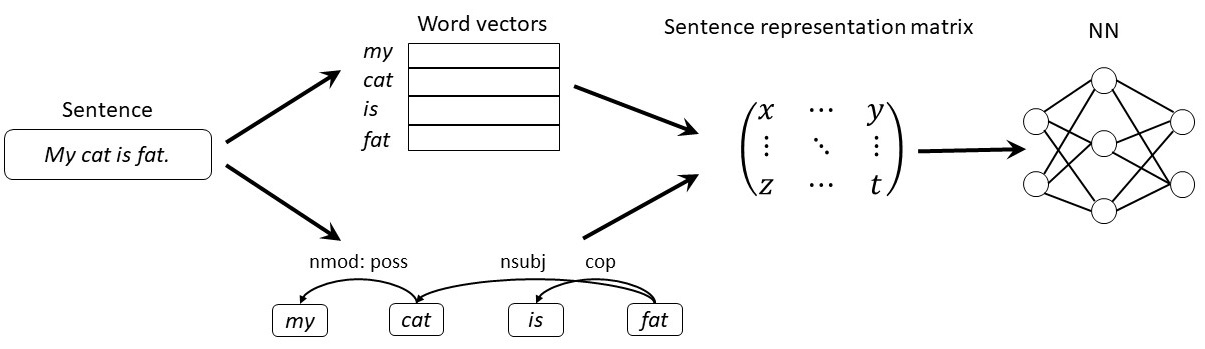}
\vspace{4mm}
\caption{\label{pipeline-fig}Pipeline of our approach}
\end{figure}

Our approach uses a matrix representation of a sentence, where every word and every syntactic dependency in that sentence is represented by a vector. Figure \ref{pipeline-fig} depicts our pipeline. 

We define several deep neural network architectures that combine CNN and LSTM layers in a different way.
We train every network on preprocessed\footnote{We applied the following text preprocessing steps: sentence boundary detection, tokenization, and dependency parsing with Stanford CoreNLP package~\citep{manning2014stanford}.} text data, where every sentence is labeled as a definition or a non-definition.
During testing, we use the pre-trained network for DE.

\subsection{Sentence representation parameters}
To represent \textit{sentence words}, we use standard sentence modeling for CNNs\\ \citep{kim2014convolutional}, where every word $w$ is represented by its $k$-dimensional word vector $\vec{w}$~\citep{mikolov2013distributed}, 
and all sentences
are assumed as having the same length $n$, using zero padding where necessary. An entire sentence is then represented by $n\times k$ zero-padded matrix $S_{n\times k}$.
In all cases we used word vectors of size $k=300$ and $n=\max_{i} \{\mbox{length of sentence \#i}\}$. 

For BERT sentence representation~\citep{devlin2018bert}, we obtain a vector of size 1024 produced by a BERT model for every sentence. The model we use is a pre-trained 24-layer uncased BERT model	with 1024-hidden layers, 16-heads, and 340M parameters released by Google AI~\citep{devlin2018bert}. 
In this case, sentence length has no effect on data representation.

Syntactic dependency, in the dependency parse tree of a sentence, has the form
$(w_{i},w_{j},\mathit{d_{ij}})$, where $w_{i},w_{j}$ are words and $\mathit{d_{ij}}$ is the dependency label.\footnote{The Stanford CoreNLP parser supports 46 dependency types.} For example, in a sentence \enquote*{\textit{This time around, they're moving even faster},} a tuple $(\mathit{moving},\mathit{they}, \mathit{nsubj})$ represents the dependency named $\mathit{nsubj}$, which connects the word $\mathit{moving}$ to the word $\mathit{they}$.

We represent dependency words $w_{i},w_{j}$ of dependency $(w_{i},w_{j},\mathit{d_{ij}})$ by a single vector, denoted by $\vec{r_{ij}}$, computed in one of the following ways:
\begin{itemize}
    \item Normalized sum $\vec{r_{ij}}^{avg}:=\frac{1}{2}(\vec{w_{i}}+\vec{w_{j}})$ of word vectors $\vec{w_{i}}$ and $\vec{w_{j}}$.
    The resulting vector has 300 dimensions.
    \item Concatenation $\vec{r_{ij}}^{c}:=\vec{w_{i}}\circ \vec{w_{j}}$ of the corresponding word vectors $\vec{w_{i}}$ and $\vec{w_{j}}$. 
    The resulting vector has 600 dimensions. 
\end{itemize} 

The dependency label $\mathit{d_{ij}}$ is represented in one of the following ways:
\begin{itemize}
    \item One-hot representation $\mathit{dep_{ij}}$ of the dependency label over the search space of size 46.
    \item Concatenation $\mathit{dep_{ij}}\circ \mathit{depth_{ij}}$ of one-hot dependency label representation with the depth vector $\mathit{depth_{ij}}$ containing one number---the depth $n \in \mathbb{Z}^{\geq 0}$ of the dependency edge in the dependency tree (edges starting in the root of the tree have depth 0). 
\end{itemize}

\subsection{Neural network models}
In this section we describe different neural network models we have implemented and tested in this work. 

\subsubsection{Neural network structures}
\begin{figure}[!th] 
\center
\includegraphics[scale=0.45]{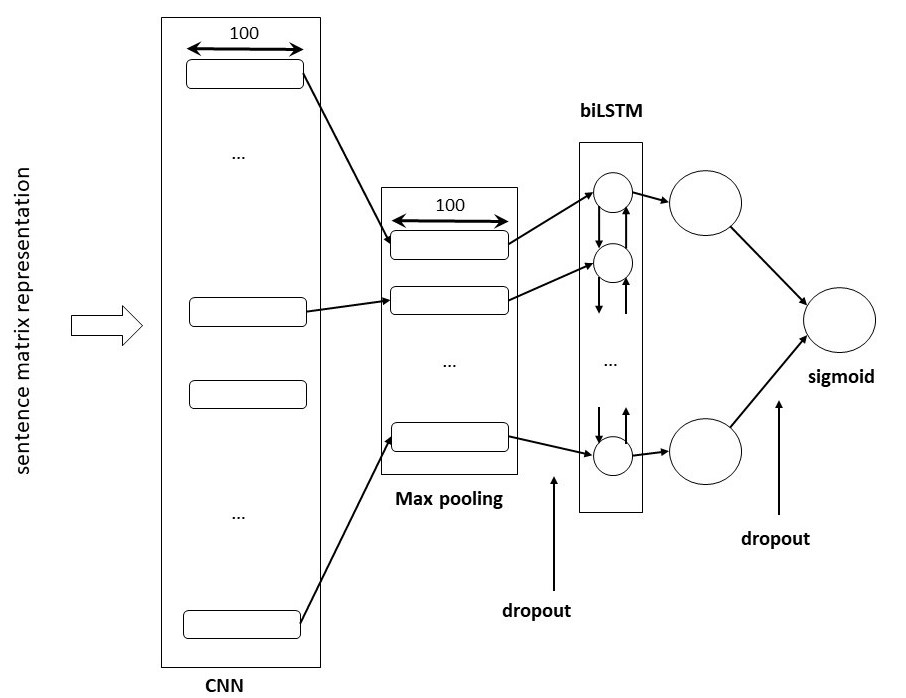}
\vspace{5mm}
\caption{\label{CBLSTM}CBLSTM network architecture.}
\end{figure}

\begin{figure}[!th]
\center
\includegraphics[scale=0.45]{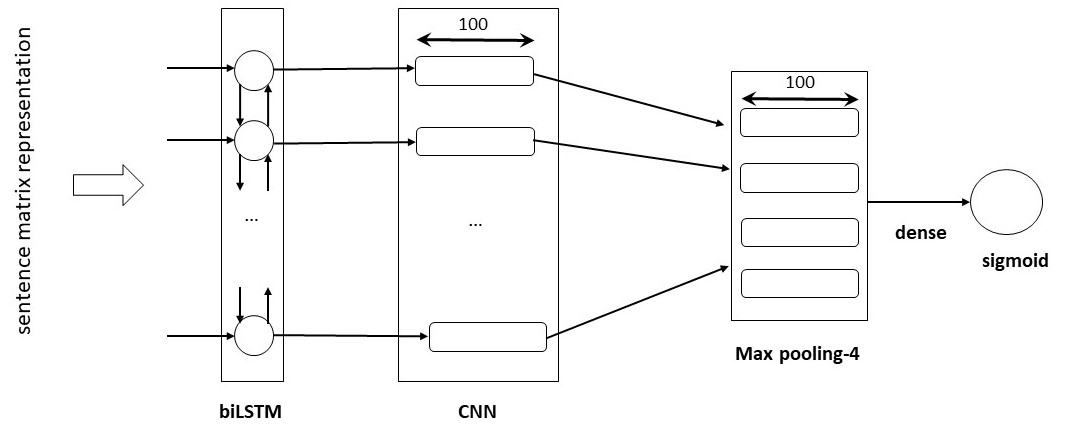}
\vspace{5mm}
\caption{\label{BLSTMCNN}BLSTMCNN network architecture.}
\end{figure}

We use four different network configurations, described below:
\begin{enumerate}
\item The convolutional network, denoted by $\mathit{CNN}$, uses a convolutional layer~\citep{lecun1998gradient} only.
\item The $\mathit{CBLSTM}$ network uses a CNN layer followed by a bidirectional LSTM layer, following the approach of \cite{anke2018syntactically}. 
\item The recurrent network that uses a single LSTM layer, denoted by $\mathit{LSTM}$.
\item The $\mathit{BLSTMCNN}$ network that uses a bidirectional LSTM layer followed by a CNN layer.
\end{enumerate}
Figure~\ref{CBLSTM} and Figure~\ref{BLSTMCNN} demonstrate mixed NN architectures---CBLSTM and BLSTMCNN, respectively. 

LSTM, CNN, and CBLSTM were tested as baselines, used in~\citep{anke2018syntactically}. Also, the inverse combination of layers, denoted by BLSTMCNN, was used  to test our hypothesis that automatically extracted features provide a better representation model for classified sentences. CNN was added as a layer that can automatically extract features and LSTM as a classification model that is context-aware. The experiment using different orders of these layers was aimed to examine which order is beneficial for the DE task---first to extract features from the original input and then feed them to the context-aware classifier, or first to calculate hidden states with context-aware LSTM gates and then feed them into CNN classifier (which includes feature extraction before the classification layer). 




\subsubsection{\label{input-sec}Input representation}
\begin{figure}[!th]
\center
\includegraphics[scale=0.6]{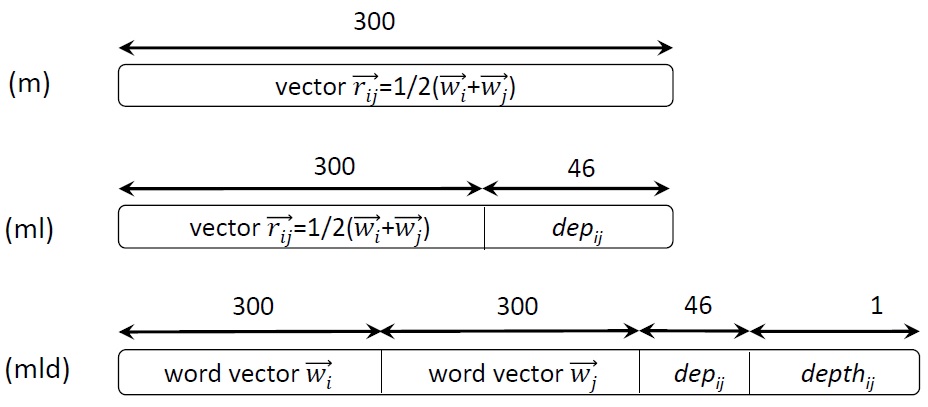}
\vspace{4mm}
\caption{\label{input-fig}Input configurations}
\end{figure}

The final sentence representation is a concatenation of the word vector matrix with the dependency structure representation, enriched with the dependency label information. 
Below, we outline three main input configurations that we have defined and tested on all our networks.
\begin{enumerate}
\item Configuration $\mathit{m}$ includes 
word vectors for sentence words  
and the words of dependencies (normalized sum of word vectors). Formally, $\mathit{m}=S_{n\times k}\circ [\vec{r_{ij}}^{avg}]_{ij}$
\item Configuration $\mathit{ml}$ includes word vectors for sentence words, 
dependency words, and dependency label representations.
 Formally, $\mathit{ml}=S_{n\times k}\circ [\vec{r_{ij}}^{avg}\circ \mathit{dep_{ij}}]_{ij}$
\item Configuration $\mathit{mld}$ has full dependency information, including concatenation of word vectors for dependency words, dependency label, and dependency depth. 
Formally, $\mathit{mld}=[\vec{r_{ij}}^{c}\circ \mathit{dep_{ij}}\circ \mathit{depth_{ij}}]_{ij}$
\end{enumerate}
Figure \ref{input-fig} shows how dependencies are represented for different configurations ($\mathit{m}$, $\mathit{ml}$, and $\mathit{mld}$, respectively). 

\begin{figure}[t!]
  \centering
       \includegraphics[scale=0.4]{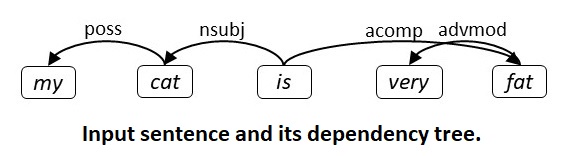}
       \\
        \includegraphics[scale=0.45]{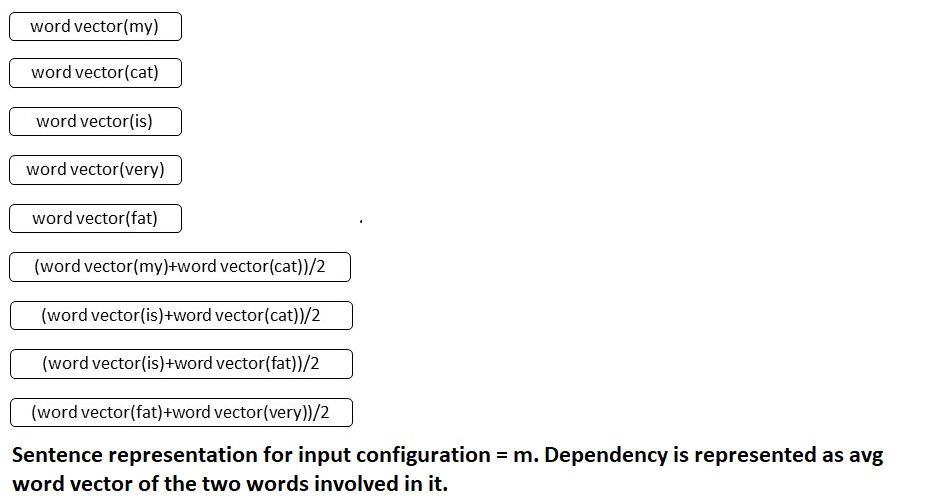}
        \includegraphics[scale=0.45]{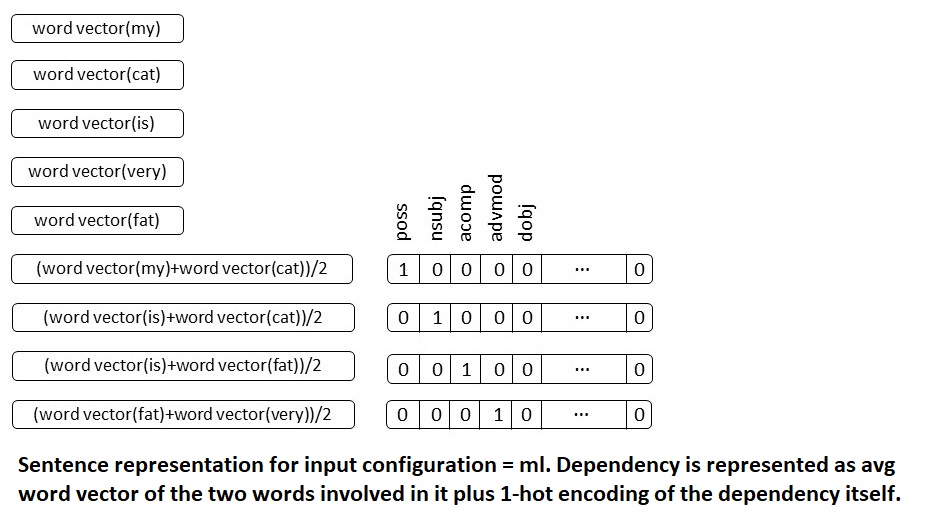}
        \includegraphics[scale=0.45]{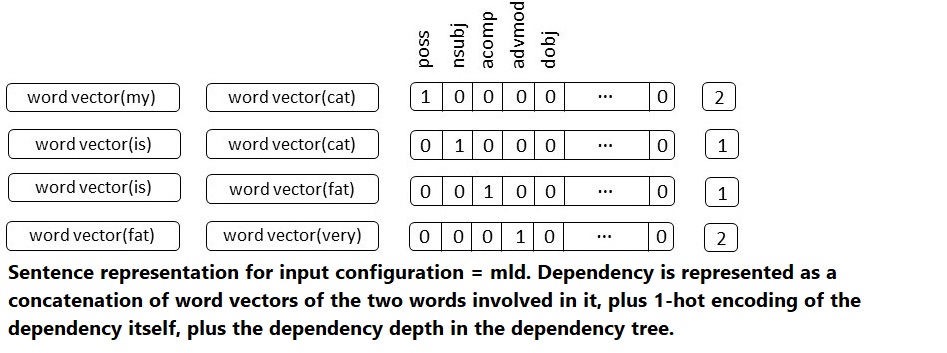}
   \vspace{2mm}
    \caption{Examples of input configurations.}\label{conf-ex}
\end{figure}

Figure~\ref{conf-ex} shows the example of these input  configurations. 


\section{Experiments}\label{ev-sec}

Our experiments aim at testing the following hypotheses:
\begin{enumerate}
    \item Deep NNs outperform classical machine learning models on DE task;
    \item CNN layer improves NN performance on DE task;
    \item Syntactic (dependency) knowledge improves  NN performance on DE task;
    \item FastText embedding performs better on our datasets than other embedding models, due to larger word coverage (because a larger portion of words from the datasets is contained in the fastText dictionary);
    \item Self-embedding performs worst among tested embeddings due to a smaller amount of training data;
    \item Because mathematical and general definitions are different, supervised DE tasks for mathematical definitions must be trained on mathematical domains;
    \item Mathematical Wikipedia articles can be automatically detected as definition-containing articles;
\end{enumerate}

Tests were performed on a cloud server with 32GB of RAM, 150 GB of PAGE memory,  an Intel Core I7-7500U 2.70 GHz CPU, and two NVIDIA GK210GL GPUs.
\subsection{Tools}
The models were implemented with help of the following tools: (1) Stanford CoreNLP wrapper~\citep{stanfordnlp-python} 
for Python (tokenization, sentence boundary detection, and dependency parsing) , (2) gensim~\citep{rehurek_lrec} (loading word2vec vectors), (3) Keras~\citep{chollet2015keras} with Tensorflow~\citep{tensorflow2015-whitepaper}  
as a back-end (NN models), 
(4) fastText vectors pre-trained on English webcrawl and Wikipedia~\citep{fasttext}, (5) Scikit-Learn~\citep{scikit-learn} (evaluation with F1, recall, and precision metrics), (6) BERT as a service python package~\citep{bert-service}, and (7) WEKA software~\citep{hall2009weka}.
All networks were trained with batch size 32 and 10 epochs.
\subsection{Data}
In our work we use the three datasets--WCL, W00 and WFMALL--that are described below.
For in-domain tests, every dataset was evaluated on its own. 
In cross-domain tests, a network or a baseline was trained on one of the datasets and was tested on another. 
Additionally,
we used a joint multi-domain dataset that contains the mathematical WFMALL dataset with the other two datasets (WCL\&W00\&WFMALL), denoted MULTI.
The number of sentences for each class, majority vote values, total number of words, and number of joint words with three pre-trained word embedding models ---word2vec (denoted by W2V), fastText (denoted by FT), and BERT---are given in Table~\ref{dataset-table}.

\begin{table*}
\scriptsize
\begin{tabular}{llllllll}
\hline\hline
Dataset	&	Definition	&	Non-def	&	Majority & Total & Common & Common & Common\\
	& sentences		&	sentences  & vote &words&  W2V&  FT&  BERT\\
\hline
WCL	&	1,871	&	2,847	&	0.603 & 21,843 & 14,368 &	16,937& 10,740 \\
W00	&	731	&	1,454	&	0.665	& 7,478 & 5,307 & 6,077 & 4,329\\
WFMALL	&	1,934	&	4,206	&	0.685	&  9,759 & 6,052 & 7,366 &  5,025\\
MULTI &	4,536	&	8,507	&	0.652	& 30,791 & 18,037  & 22,155 & 12,678 \\
\hline
\end{tabular}
\vspace{1mm}
\caption{Datasets description}\label{dataset-table}
\end{table*}

\subsubsection{The WCL Dataset}
The World-Class Lattices (WCL) dataset~\citep{WCL}, introduced in \citep{navigli2010annotated}, was constructed from manually annotated Wikipedia data
in English. The version that we used (WCL v1.2) contains 4,719 annotated sentences, 1,871 of which are proper definitions and 2,847 are distractor sentences, that
have similar structures with proper definitions, but are not actually definitions. This dataset contains generic definitions in all areas and it is not
mathematically oriented. A sample definition sentence from this dataset is
$$\begin{array}{l}\mbox{\it American Standard Code for Information Interchange}\\ 
\mbox{\it is a character encoding based on English alphabet.}\end{array}$$
and a sample distractor is
$$\begin{array}{l}\mbox{\it The premise of the program revolves around Parker,}\\ 
                  \mbox{\it an 18-year-old country girl who moves back}\\
				  \mbox{\it  and forth between her country family, who lives }\\
                  \mbox{\it on a bayou houseboat, and the wealthy Brents, who own}\\
                  \mbox{\it a plantation and pancake business.}\\
\end{array}$$
In this corpus, following parts of definitions are annotated: 
\begin{enumerate}
\item the DEFINIENDUM field (DF), referring to the word being defined and its modifiers, 
\item the DEFINITOR field (VF),
referring to the verb phrase used to introduce a definition,
\item the DEFINIENS field (GF), which includes the genus phrase, and \item the REST field (RF), which indicates all additional sentence parts.
\end{enumerate}
According to the original annotations, existence of the first three parts indicates that a sentence is a definition. 

\subsubsection{The W00 Dataset}
The W00 dataset~\citep{W00}, introduced in \citep{jin2013mining}, 
was compiled from ACL-ARC ontology~\citep{bird2008acl} and contains 2,185 manually annotated sentences,
with 731 definitions and 1,454 non-definitions; the style of the distractors is different from the one used in the WCL dataset.
A sample definition sentence from this dataset is
$$\begin{array}{l}\mbox{\it Our system, SNS (pronounced “essence”), retrieves}\\
                  \mbox{\it documents to an unrestricted user query and }\\
                  \mbox{\it summarizes a subset of them as selected by the user.}\end{array}$$
and a sample distractor is
$$\begin{array}{l}\mbox{\it The senses with the highest confidence scores are the}\\
\mbox{\it senses that contribute the most to the function for the set.
}
\end{array}$$
Annotation of the W00 dataset is token-based, with each token in a sentence identified by a single label that indicates whether a token is a part of a term ($T$), a definition ($D$), or neither ($O$). According to the original annotation, a sentence is considered not to be a definition if all of its tokens are marked as $O$. Sentence that contains tokens marked as $T$ or $D$ is considered to be a definition.

\subsubsection{The WFMALL Dataset}
The WFMALL dataset is an extension of the WFM dataset~\citep{WFM}. It was created by us after collecting and processing all 2,352 articles from Wolfram Mathworld~\citep{weisstein2007wolfram}. The final dataset contains 6,140 sentences, of which 1,934 are definitions and 4,206 are non-definitions.
Sentences were extracted automatically and then manually separated into two categories: definitions and statements (non-definitions).
All annotators (five in total) have at least BSc degree and learned academic mathematical courses (research group members, including three research students). The data was semi-automatically segmented to sentences with Stanford CoreNLP package and then manually assessed. All malformed sentences (as result of wrong segmentation) were fixed, 116 too short sentences (with less than 3 words) were removed. All sentences related to Wolfram Language\footnote{\url{https://en.wikipedia.org/wiki/Wolfram_Language}} were removed because they relate to a programming language and describe how mathematical objects are expressed in this language, and not how they are defined. Sentences with formulas only, without text, were also removed. The final dataset was split to nine portions, saved as Unicode text files. Three annotators worked on each portion. 
First, two annotators labeled sentences independently. Then, all sentences that were given different labels were finally annotated by the third annotator  (controller)\footnote{We decided that a  label with majority vote will be selected. Therefore, the third annotator (controller) labeled only the sentences with contradict labels.}. The final label was set by majority vote. 
The kappa agreement between annotators was 0.651, which is considered substantial agreement. 

This dataset is freely available for download from GitHub.\footnote{\url{https://drive.google.com/drive/folders/1052akYuxgc2kbHH8tkMw4ikBFafIW0tK?usp=sharing}}
A sample definition sentence from this dataset is
$$\begin{array}{l}\mbox{\it The $(7,3,2)$-von Dyck group, also sometimes termed the}\\
                  \mbox{\it $(2,3,7)$-von Dyck group, is defined as the von Dyck group}\\
									\mbox{\it with parameters $(7,3,2)$.}\end{array}$$
and a sample non-definition is
$$\begin{array}{l}\mbox{\it Any $2$-Engel group must be a group of nilpotency class $3$.}
\end{array}$$

\subsection{\label{pre-sec}Text preprocessing}
With regard to all three datasets described 
above, we applied the same text preprocessing steps in the following manner:
\begin{itemize}
\item Sentence splitting was derived explicitly from the datasets, without applying any additional procedure, in the following manner: WCL and W00 datasets came pre-split, and sentence splitting for the new WFMALL dataset was performed semi-automatically by our team (using Stanford CoreNLP SBD, followed by manual correction, due to many formulas in the text).
\item Tokenization and dependency parsing were performed on all sentences with the help of the Stanford CoreNLP package~\citep{manning2014stanford}.
\item For the WCL and W00 datasets used in \cite{anke2018syntactically} for DE, we replaced parsing by SpaCy~\citep{honnibal-johnson:2015:EMNLP}  with the Stanford CoreNLP parser. We found that the latter tool improved precision. 
\item We have tested three pre-trained word embedding options, the word2vec~\citep{word2vec} with pre-trained  vectors obtained from the Google News corpus, fastText~\citep{grave2018learning} with vectors pre-trained on English webcrawl and Wikipedia (available at \citep{fasttext}), and BERT~\citep{devlin2018bert}. Also, self-embedding vectors that were trained on our data with an embedding layer, were tested and compared to the pre-trained models. 
\end{itemize}

\subsection{Baselines}
Using the WEKA software~\citep{hall2009weka}, we applied the following baselines: Simple Logistic Regression (SLR), Support Vector Machine (SVM) and Random Forest (RF).
To use these methods on complete sentences, we computed vector representation for every sentence as an average of word vectors for all its words, and marked sentences as belonging to either a `definition' or a `non-definition'  class.
We have also used the DefMiner system of~\cite{jin2013mining}, available at \url{https://github.com/YipingNUS/DefMiner}, as a baseline. Because DefMiner comes pre-trained, we do not use it in the cross-domain evaluations. 

\subsection{\label{res-sec}Results}
We tested all network configurations and baselines (where applicable) in three domain configurations described in the following sections. We use the notation $\mathit{Network}_{\mathit{input}}$ for a network of type $\mathit{Network}$, accepting input of type $\mathit{input}$ as specified in Section \ref{input-sec}.

In total, we have four network types, three input configurations, three word embedding models for each network, and pre-trained BERT embedding that resulted in four additional configurations (one per network, because we did not combine it with dependency features), resulting in 40 final configurations. Our motivation was to test the extent to which the order and presence of CNN and LSTM layers affect the results, and how the performance was affected by the representation of sentence dependency information and different embedding models.\footnote{Our python code is available at \url{https://github.com/NataliaVanetik/MDE-v2}.}
In the tables containing results, we show accuracy and F1 scores for all embeddings, with the best scores for each dataset marked in bold.

The first series of tests was performed for every dataset 
separately. 
Each dataset was divided, by a stratified sampling, into three sets: training=70\%, validation=5\%, test=25\%. Every model was trained on a training set and tested on a test set. We adjusted the hyperparameters of all NN models using a validation set.
\subsubsection{In-domain and multi-domain results}
Results of in-domain experiments are given in Tables~\ref{in-domain-tab-WCL} through \ref{in-domain-tab-WFM}.
and visualized in Figures~\ref{in-domain-wcl} through \ref{in-domain-wfm} (see Appendix).  
Because DefMiner does not work with embedding vectors, its scores for all embedding models are identical. 
\begin{table}
\scriptsize
\begin{tabular}{lllllllll}
\hline				
										
Method	&	W2V	&	FT	&	BERT    &   self    &   W2V	&	FT	&	BERT    &   self        \\
	&	acc	&	acc	& acc	&	acc	&	F1	&	F1	&	F1	&	F1 	\\
\hline												
RF	&	0.779	&	0.807	&	0.878	&	0.721	&	0.773	&	0.800	&	0.875	&	0.600	\\
SL	&	0.817	&	0.859	&	\textbf{0.935}	&	0.775	&	0.817	&	0.859	&	0.934	&	0.771	\\
SVM	&	0.824	&	0.756	&	0.934	&	0.780	&	0.824	&	0.863	&	\textbf{0.936}	&	0.776	\\
DefMiner	&	0.797	&	0.797	&	0.797	&	0.797	&	0.741	&	0.741	&	0.741	&	0.741	\\
\hline					&		&						&		&		\\
$\mathrm{CNN_{m}}$	&	0.889	&	0.916	&	0.925	&	0.909	&	0.889	&	0.916	&	0.925	&	0.908	\\
$\mathrm{CNN_{ml}}$	&	0.909	&	0.930	&	0.925	&	0.934	&	0.909	&	0.930	&	0.925	&	0.934	\\
$\mathrm{CNN_{mld}}$	&	0.915	&	0.929	&	0.925	&	0.917	&	0.915	&	0.929	&	0.925	&	0.917	\\
$\mathrm{CBLSTM_{m}}$	&	0.892	&	0.903	&	0.825	&	0.931	&	0.891	&	0.902	&	0.824	&	0.931	\\
$\mathrm{CBLSTM_{ml}}$	&	0.921	&	0.929	&	0.825	&	\textbf{0.935}	&	0.921	&	0.929	&	0.824	&	\textbf{0.935}	\\
$\mathrm{CBLSTM_{mld}}$	&	0.914	&	0.926	&	0.825	&	0.916	&	0.914	&	0.926	&	0.824	&	0.917	\\
$\mathrm{LSTM_{m}}$	&	0.858	&	0.871	&	0.618	&	0.913	&	0.858	&	0.871	&	0.506	&	0.914	\\
$\mathrm{LSTM_{ml}}$	&	0.876	&	0.890	&	0.618	&	0.914	&	0.876	&	0.889	&	0.506	&	0.914	\\
$\mathrm{LSTM_{mld}}$	&	0.905	&	0.919	&	0.618	&	0.933	&	0.905	&	0.920	&	0.506	&	0.933	\\
$\mathrm{BLSTMCNN_{m}}$	&	0.889	&	0.912	&	0.922	&	0.915	&	0.889	&	0.912	&	0.922	&	0.915	\\
$\mathrm{BLSTMCNN_{ml}}$	&	\textbf{0.923}	&	0.916	&	0.922	&	0.913	&	\textbf{0.927}	&	0.915	&	0.922	&	0.912	\\
$\mathrm{BLSTMCNN_{mld}}$	&	0.919	&	\textbf{0.933}	&	0.922	&	0.913	&	0.920	&	\textbf{0.933}	&	0.922	&	0.913	\\
\hline																						\end{tabular}
\vspace{1mm}
\caption{In-domain performance for WCL dataset.}\label{in-domain-tab-WCL}
\end{table}

\begin{table}
\scriptsize
\begin{tabular}{lllllllll}
\hline				
										
Method	&	W2V	&	FT	&	BERT    &   self    &   W2V	&	FT	&	BERT    &   self        \\
	&	acc	&	acc	& acc	&	acc	&	F1	&	F1	&	F1	&	F1 	\\
\hline												
RF	&	0.717	&	0.705	&	0.702	&	0.678	&	0.659	&	0.637	&	0.622	&	0.591	\\
SL	&	0.703	&	0.721	&	0.722	&	0.690	&	0.677	&	0.697	&	0.716	&	0.640	\\
SVM	&	0.707	&	0.717	&	0.701	&	0.682	&	0.645	&	0.665	&	0.700	&	0.575	\\
DefMiner	&	\textbf{0.819}	&	\textbf{0.819}	&	\textbf{0.819}	&	\textbf{0.819}	&	0.644	&	0.644	&	0.644	&	0.644	\\
\hline	 		 		 		 		 				 		 		
$\mathrm{CNN_{m}}$	&	0.696	&	0.698	&	0.769	&	0.652	&	0.622	&	0.607	&	\textbf{0.766}	&	0.636	\\
$\mathrm{CNN_{ml}}$	&	0.691	&	0.698	&	0.769	&	0.714	&	0.617	&	0.610	&	\textbf{0.766}	&	0.668	\\
$\mathrm{CNN_{mld}}$	&	0.698	&	0.700	&	0.769	&	0.696	&	0.631	&	0.617	&	\textbf{0.766}	&	0.685	\\
$\mathrm{CBLSTM_{m}}$	&	0.696	&	0.709	&	0.684	&	0.698	&	0.656	&	0.677	&	0.556	&	0.696	\\
$\mathrm{CBLSTM_{ml}}$	&	0.712	&	0.716	&	0.684	&	0.732	&	0.676	&	0.668	&	0.556	&	\textbf{0.707}	\\
$\mathrm{CBLSTM_{mld}}$	&	0.735	&	0.705	&	0.684	&	0.707	&	0.732	&	0.698	&	0.556	&	0.702	\\
$\mathrm{LSTM_{m}}$	&	0.684	&	0.698	&	0.684	&	0.686	&	0.649	&	0.679	&	0.556	&	0.654	\\
$\mathrm{LSTM_{ml}}$	&	0.689	&	0.696	&	0.684	&	0.675	&	0.649	&	0.643	&	0.556	&	0.618	\\
$\mathrm{LSTM_{mld}}$	&	0.723	&	0.730	&	0.684	&	0.698	&	0.706	&	\textbf{0.719}	&	0.556	&	0.697	\\
$\mathrm{BLSTMCNN_{m}}$	&	0.712	&	0.728	&	0.769	&	0.657	&	0.689	&	0.700	&	\textbf{0.766}	&	0.660	\\
$\mathrm{BLSTMCNN_{ml}}$	&	0.700	&	0.716	&	0.769	&	0.705	&	0.706	&	0.707	&	\textbf{0.766}	&	0.701	\\
$\mathrm{BLSTMCNN_{mld}}$	&	0.719	&	0.696	&	0.769	&	0.689	&	0.721	&	0.691	&	\textbf{0.766}	&	0.683	\\
\hline												\end{tabular}
\vspace{1mm}
\caption{In-domain performance for W00 dataset.}\label{in-domain-tab-W00}
\end{table}

\begin{table}
\scriptsize
\begin{tabular}{lllllllll}
\hline				
										
Method	&	W2V	&	FT	&	BERT    &   self    &   W2V	&	FT	&	BERT    &   self        \\
	&	acc	&	acc	& acc	&	acc	&	F1	&	F1	&	F1	&	F1 	\\
\hline												
RF	&	0.720	&	0.766	&	0.776	&	0.705	&	0.665	&	0.733	&	0.623	&	0.740	\\
SL	&	0.756	&	0.788	&	0.841	&	0.721	&	0.737	&	0.780	&	0.660	&	0.840	\\
SVM	&	0.759	&	0.792	&	\textbf{0.844}	&	0.700	&	0.736	&	0.780	&	0.700	&	\textbf{0.844}	\\
DefMiner	&	0.704	&	0.704	&	0.704	&	0.704	&	0.134	&	0.134	&	0.134	&	0.134	\\
\hline													
$\mathrm{CNN_{m}}$	&	0.834	&	0.856	&	0.827	&	0.717	&	0.830	&	0.856	&	0.821	&	0.679	\\
$\mathrm{CNN_{ml}}$	&	0.832	&	0.859	&	0.827	&	0.780	&	0.831	&	0.858	&	0.821	&	0.781	\\
$\mathrm{CNN_{mld}}$	&	\textbf{0.841}	&	\textbf{0.867}	&	0.827	&	0.785	&	0.836	&	\textbf{0.866}	&	0.821	&	0.782	\\
$\mathrm{CBLSTM_{m}}$	&	0.835	&	0.860	&	0.702	&	0.751	&	0.836	&	0.861	&	0.646	&	0.752	\\
$\mathrm{CBLSTM_{ml}}$	&	0.840	&	0.864	&	0.702	&	0.744	&	0.835	&	0.865	&	0.646	&	0.742	\\
$\mathrm{CBLSTM_{mld}}$	&	0.835	&	0.855	&	0.702	&	0.789	&	0.831	&	0.857	&	0.646	&	0.784	\\
$\mathrm{LSTM_{m}}$	&	0.832	&	0.853	&	0.673	&	0.742	&	0.829	&	0.847	&	0.576	&	0.731	\\
$\mathrm{LSTM_{ml}}$	&	0.828	&	0.860	&	0.673	&	0.746	&	0.828	&	0.858	&	0.576	&	0.734	\\
$\mathrm{LSTM_{mld}}$	&	\textbf{0.841}	&	0.849	&	0.673	&	\textbf{0.801}	&	\textbf{0.839}	&	0.850	&	0.576	&	0.801	\\
$\mathrm{BLSTMCNN_{m}}$	&	0.829	&	0.849	&	0.833	&	0.751	&	0.823	&	0.850	&	\textbf{0.831}	&	0.748	\\
$\mathrm{BLSTMCNN_{ml}}$	&	0.833	&	0.851	&	0.833	&	0.781	&	0.831	&	0.849	&	\textbf{0.831}	&	0.778	\\
$\mathrm{BLSTMCNN_{mld}}$	&	0.828	&	0.859	&	0.833	&	0.779	&	0.828	&	0.858	&	\textbf{0.831}	&	0.782	\\
\hline												\end{tabular}
\vspace{1mm}
\caption{In-domain performance for WFMALL dataset.}\label{in-domain-tab-WFM}
\end{table}

We decided not to merge  BERT representation with dependency knowledge because: (1) BERT did not provide any performance advantage over other embedding models without dependency knowledge; (2) BERT representation produces long  vectors (1024  dimensions in our case) that require a large amount of memory, and the BERT-as-a-service package takes an extended period of time to compute sentence vectors, which significantly slows the classification task, making it almost impossible to run all combinations of input configurations and networks within available time constraints. Following this decision, the scores for BERT embedding do not depend on the representation and appear to be the same for all three ($m$, $ml$, and $mld$).

The first observation that can be seen from the 
results is that all NN-based models usually perform much better than four baselines. From Tables~\ref{in-domain-tab-WCL}  through \ref{in-domain-tab-WFM} it can be seen that models having a CNN layer as one (or only one) of their components outperform other models in most cases. Also, results demonstrate that dependency knowledge improves the performance of the models.  

It is worth noting that the DefMiner on W00 superiority with accuracy scores can be naturally  explained by the fact that DefMiner was designed  by extracting hand-crafted shallow parsing patterns from the W00 dataset. DefMiner gives very poor F1 scores on the WFMALL dataset because it almost always classifies mathematical definitions as a non-definition. \footnote{Confusion matrix values are TP=130,	FN=1804, TN=4195 and FP=12. That gave us high precision P=0.915 and low recall R1=0.072, resulting in a low F1 value.}

Another interesting phenomenon was observed -- there are cases where baselines performed better than NN models. All these cases have something in common: the  sentences  were   represented  with  BERT  or  self-embedding  vectors. 

As such, the following general conclusions can be made: (1) the CNN model---pure or integrated with BLSTM---achieves better performance than LSTM; (2) the NN models gain better performance with dependency information. 
The superiority of models having a CNN layer  
can be explained by the ability of CNN to learn features and reduce the number of free parameters in a high-dimensional  sentence representation, allowing the network to be more accurate with fewer parameters. Due to a high-dimensional input in our task, this characteristic of CNN appears to be very helpful. 

\begin{table}
\scriptsize
\begin{tabular}{lllllllll}
\hline				
										
Method	&	W2V	&	FT	&	BERT    &   self    &   W2V	&	FT	&	BERT    &   self        \\
	&	acc	&	acc	& acc	&	acc	&	F1	&	F1	&	F1	&	F1 	\\
\hline												
RF	&	0.726	&	0.750	&	0.777	&	0.695	&	0.690	&	0.722	&	0.752	&	0.626	\\
SL	&	0.749	&	0.784	&	0.841	&	0.711	&	0.734	&	0.775	&	0.840	&	0.675	\\
SVM	&	0.749	&	0.776	&	\textbf{0.851}	&	0.698	&	0.727	&	0.763	&	\textbf{0.850}	&	0.636	\\
DefMiner	&	0.766	&	0.766	&	0.766	&	0.766	&	0.549	&	0.549	&	0.549	&	0.549	\\
\hline													
$\mathrm{CNN_{m}}$	&	0.839	&	0.859	&	0.843	&	0.713	&	0.838	&	0.860	&	0.837	&	0.705	\\
$\mathrm{CNN_{ml}}$	&	0.844	&	\textbf{0.870}	&	0.843	&	0.787	&	0.844	&	\textbf{0.870}	&	0.837	&	0.785	\\
$\mathrm{CNN_{mld}}$	&	0.847	&	0.859	&	0.843	&	0.774	&	0.847	&	0.860	&	0.837	&	0.775	\\
$\mathrm{CBLSTM_{m}}$	&	0.841	&	\textbf{0.870}	&	0.765	&	0.732	&	0.840	&	0.868	&	0.761	&	0.735	\\
$\mathrm{CBLSTM_{ml}}$	&	0.850	&	0.861	&	0.765	&	0.732	&	0.851	&	0.857	&	0.761	&	0.735	\\
$\mathrm{CBLSTM_{mld}}$	&	0.841	&	0.865	&	0.765	&	0.785	&	0.842	&	0.864	&	0.761	&	0.782	\\
$\mathrm{LSTM_{m}}$	&	\textbf{0.856}	&	0.865	&	0.667	&	0.752	&	\textbf{0.853}	&	0.863	&	0.572	&	0.750	\\
$\mathrm{LSTM_{ml}}$	&	0.796	&	0.861	&	0.667	&	0.617	&	0.790	&	0.860	&	0.572	&	0.569	\\
$\mathrm{LSTM_{mld}}$	&	0.851	&	0.861	&	0.667	&	0.785	&	0.850	&	0.861	&	0.572	&	0.781	\\
$\mathrm{BLSTMCNN_{m}}$	&	0.853	&	0.849	&	0.829	&	0.749	&	\textbf{0.853}	&	0.847	&	0.818	&	0.746	\\
$\mathrm{BLSTMCNN_{ml}}$	&	0.844	&	0.860	&	0.829	&	\textbf{0.800}	&	0.843	&	0.859	&	0.818	&	\textbf{0.798}	\\
$\mathrm{BLSTMCNN_{mld}}$	&	0.842	&	0.864	&	0.829	&	0.787	&	0.840	&	0.863	&	0.818	&	0.783	\\
\hline												\end{tabular}
\vspace{1mm}
\caption{Multi-domain performance.}\label{in-domain-tab-MULTI}
\end{table}

The second series of tests was performed on the joint dataset \\
WCL\&W00\&WFMALL (denoted by MULTI). The dataset was randomly split to 70\%-25\%-5\% training, test, and validation data, respectively (keeping the proportions of mixed data). Results are given in  Table~\ref{in-domain-tab-MULTI} and in  Figure~\ref{in-domain-multi} (in Appendix). 

Table~\ref{in-domain-tab-MULTI} also shows that the CNN layer improves the performance of our NN models. However, it does not demonstrate consistent superiority for  incorporating syntactic information into sentence representations.  
%

Statistical analysis performed on the obtained scores shows the following:
(1) On two datasets (WFMALL and MULTI) there is a significant superiority of two pre-trained word embeddings (word2vec and fastText) over the other two embeddings (self-trained and pre-trained BERT); there is no significant difference between the four embedding models on the other two datasets; 
(2) FastText is significantly better than most other representations on two datasets; 
(3) There is a significant superiority of CNN and CBLSTM models over LSTM in many configurations (WCL using m and ml, WFMALL using m; W00 and MULTI using ml); 
(4) There is a significant improvement when we use dependency information in the sentence representation (ml and mld vs m) in many cases (WCL: CNN, CBLSTM, LSTM; WFMALL: CNN; W00: LSTM); 
(5) Some evaluated models on three datasets (WCL: CBLSTM and CNN using ml and mld; WFMALL: CNN using mld; MULTI: CNN with ml and mld) are significantly better than the best baseline (SVM).

In conclusion, we would recommend to use CNN or CBLSTM models with fastText embedding and ml or mld input configuration for DE in both general and mathematical domains. 

\begin{table}
\scriptsize
\begin{tabular}{lllllllllllllllll}
\hline									
Method	&	W2V	&	FT	&	BERT & self & W2V	&	FT	&	BERT & self \\
	&	acc	&	acc	&	acc	&	acc	& F1	&	F1	& F1	&	F1	\\
\hline													
RF	&	0.638	&	0.642	&	0.638	&	0.599	&	0.549	&	0.547	&	0.529	&	0.459	\\
SL	&	0.688	&	0.751	&	\textbf{0.864}	&	0.601	&	0.674	&	0.746	&	\textbf{0.861}	&	0.464	\\
SVM	&	0.681	&	0.745	&	0.841	&	0.604	&	0.654	&	0.738	&	0.838	&	0.459	\\
\hline													
$\mathrm{CNN_{m}}$	&	0.723	&	0.723	&	0.776	&	0.604	&	0.698	&	0.698	&	0.755	&	0.455	\\
$\mathrm{CNN_{ml}}$	&	0.685	&	0.781	&	0.776	&	0.702	&	0.633	&	0.775	&	0.755	&	0.661	\\
$\mathrm{CNN_{mld}}$	&	0.669	&	0.750	&	0.776	&	0.658	&	0.608	&	0.732	&	0.755	&	0.574	\\
$\mathrm{CBLSTM_{m}}$	&	0.764	&	0.764	&	0.651	&	0.605	&	0.756	&	0.756	&	0.570	&	0.468	\\
$\mathrm{CBLSTM_{ml}}$	&	0.754	&	\textbf{0.806}	&	0.651	&	0.603	&	0.737	&	\textbf{0.796}	&	0.570	&	0.481	\\
$\mathrm{CBLSTM_{mld}}$	&	0.668	&	0.785	&	0.651	&	0.653	&	0.602	&	0.784	&	0.570	&	0.562	\\
$\mathrm{LSTM_{m}}$	&	0.772	&	0.772	&	0.608	&	0.600	&	0.758	&	0.758	&	0.469	&	0.474	\\
$\mathrm{LSTM_{ml}}$	&	\textbf{0.805}	&	0.801	&	0.608	&	0.602	&	\textbf{0.801}	&	0.793	&	0.469	&	0.477	\\
$\mathrm{LSTM_{mld}}$	&	0.693	&	0.737	&	0.608	&	0.674	&	0.647	&	0.719	&	0.469	&	0.611	\\
$\mathrm{BLSTMCNN_{m}}$	&	0.760	&	0.760	&	0.795	&	0.605	&	0.743	&	0.743	&	0.781	&	0.548	\\
$\mathrm{BLSTMCNN_{ml}}$	&	0.720	&	0.766	&	0.795	&	\textbf{0.718}	&	0.687	&	0.760	&	0.781	&	\textbf{0.698}	\\
$\mathrm{BLSTMCNN_{mld}}$	&	0.680	&	0.730	&	0.795	&	0.709	&	0.638	&	0.709	&	0.781	&	0.680	\\
\hline
\end{tabular}
\vspace{1mm}
\caption{Cross-domain scores for WFMALL and WCL datasets as training and test sets, respectively.}\label{cross-domain-tab-wfm-wcl}
\end{table}

\begin{table}
\scriptsize
\begin{tabular}{lllllllllllllllll}
\hline									
Method	&	W2V	&	FT	&	BERT & self & W2V	&	FT	&	BERT & self \\
	&	acc	&	acc	&	acc	&	acc	& F1	&	F1	& F1	&	F1	\\
\hline													
RF	&	0.673	&	0.671	&	0.669	&	0.667	&	0.567	&	0.554	&	0.540	&	0.535	\\
SL	&	0.676	&	0.674	&	0.697	&	\textbf{0.676}	&	0.607	&	0.584	&	0.630	&	0.571	\\
SVM	&	0.677	&	0.674	&	\textbf{0.702}	&	0.668	&	0.590	&	0.574	&	\textbf{0.655}	&	0.541	\\
\hline													
$\mathrm{CNN_{m}}$	&	0.702	&	0.707	&	0.690	&	0.666	&	0.665	&	0.678	&	0.598	&	0.547	\\
$\mathrm{CNN_{ml}}$	&	0.697	&	0.693	&	0.690	&	0.644	&	0.635	&	0.653	&	0.598	&	0.610	\\
$\mathrm{CNN_{mld}}$	&	0.693	&	0.693	&	0.690	&	0.669	&	0.625	&	0.652	&	0.598	&	0.589	\\
$\mathrm{CBLSTM_{m}}$	&	0.692	&	0.708	&	0.672	&	0.657	&	0.673	&	\textbf{0.683}	&	0.573	&	0.583	\\
$\mathrm{CBLSTM_{ml}}$	&	0.706	&	\textbf{0.714}	&	0.672	&	0.618	&	0.667	&	0.676	&	0.573	&	0.592	\\
$\mathrm{CBLSTM_{mld}}$	&	0.691	&	0.667	&	0.672	&	0.661	&	0.622	&	0.658	&	0.573	&	0.588	\\
$\mathrm{LSTM_{m}}$	&	0.712	&	0.701	&	0.664	&	0.650	&	0.676	&	0.640	&	0.533	&	0.589	\\
$\mathrm{LSTM_{ml}}$	&	0.712	&	0.710	&	0.664	&	0.641	&	\textbf{0.695}	&	0.670	&	0.533	&	0.589	\\
$\mathrm{LSTM_{mld}}$	&	0.682	&	0.681	&	0.664	&	0.657	&	0.622	&	0.646	&	0.533	&	0.608	\\
$\mathrm{BLSTMCNN_{m}}$	&	\textbf{0.715}	&	0.700	&	0.699	&	0.590	&	0.679	&	0.679	&	0.631	&	0.590	\\
$\mathrm{BLSTMCNN_{ml}}$	&	0.709	&	0.700	&	0.699	&	0.650	&	0.665	&	0.657	&	0.631	&	\textbf{0.632}	\\
$\mathrm{BLSTMCNN_{mld}}$	&	0.659	&	0.684	&	0.699	&	0.648	&	0.605	&	0.633	&	0.631	&	0.624	\\
\hline
\end{tabular}
\vspace{1mm}
\caption{Cross-domain scores for WFMALL and W00 datasets as training and test sets, respectively.}\label{cross-domain-tab-wfm-w00}
\end{table}

\begin{table}
\scriptsize
\begin{tabular}{lllllllllllllllll}
\hline									
Method	&	W2V	&	FT	&	BERT & self & W2V	&	FT	&	BERT & self \\
	&	acc	&	acc	&	acc	&	acc	& F1	&	F1	& F1	&	F1	\\
\hline													
RF	&	0.669	&	0.714	&	0.745	&	0.658	&	0.668	&	0.702	&	0.737	&	0.594	\\
SL	&	0.675	&	0.704	&	\textbf{0.755}	&	0.571	&	0.677	&	0.704	&	\textbf{0.745}	&	0.576	\\
SVM	&	0.678	&	0.715	&	0.744	&	0.549	&	0.680	&	0.715	&	0.741	&	0.560	\\
\hline		
$\mathrm{CNN_{m}}$	&	0.733	&	0.760	&	0.712	&	0.636	&	0.728	&	0.751	&	0.721	&	0.587	\\
$\mathrm{CNN_{ml}}$	&	0.731	&	0.751	&	0.712	&	\textbf{0.671}	&	0.729	&	0.743	&	0.721	&	\textbf{0.621}	\\
$\mathrm{CNN_{mld}}$	&	0.709	&	0.737	&	0.712	&	0.640	&	0.713	&	0.733	&	0.721	&	0.620	\\
$\mathrm{CBLSTM_{m}}$	&	0.756	&	\textbf{0.779}	&	0.686	&	0.654	&	0.750	&	\textbf{0.768}	&	0.682	&	0.577	\\
$\mathrm{CBLSTM_{ml}}$	&	\textbf{0.758}	&	0.774	&	0.686	&	0.637	&	\textbf{0.752}	&	\textbf{0.768}	&	0.682	&	0.583	\\
$\mathrm{CBLSTM_{mld}}$	&	0.717	&	0.742	&	0.686	&	0.617	&	0.720	&	0.734	&	0.682	&	0.609	\\
$\mathrm{LSTM_{m}}$	&	0.741	&	0.724	&	0.679	&	0.539	&	0.738	&	0.730	&	0.588	&	0.552	\\
$\mathrm{LSTM_{ml}}$	&	0.752	&	0.745	&	0.679	&	0.569	&	0.734	&	0.735	&	0.588	&	0.570	\\
$\mathrm{LSTM_{mld}}$	&	0.702	&	0.749	&	0.679	&	0.625	&	0.707	&	0.742	&	0.588	&	0.616	\\
$\mathrm{BLSTMCNN_{m}}$	&	0.731	&	0.754	&	0.743	&	0.496	&	0.726	&	0.752	&	0.737	&	0.514	\\
$\mathrm{BLSTMCNN_{ml}}$	&	0.739	&	0.742	&	0.743	&	0.632	&	0.709	&	0.731	&	0.737	&	0.616	\\
$\mathrm{BLSTMCNN_{mld}}$	&	0.705	&	0.725	&	0.743	&	0.631	&	0.705	&	0.722	&	0.737	&	0.614	\\
\hline
\end{tabular}
\vspace{1mm}
\caption{Cross-domain scores for WCL and WFMALL datasets as training and test sets, respectively.}\label{cross-domain-tab-wcl-wfm}
\end{table}

\begin{table}
\scriptsize
\begin{tabular}{lllllllllllllllll}
\hline									
Method	&	W2V	&	FT	&	BERT & self & W2V	&	FT	&	BERT & self \\
	&	acc	&	acc	&	acc	&	acc	& F1	&	F1	& F1	&	F1	\\
\hline													
RF	&	0.633	&	0.658	&	0.694	&	0.657	&	0.627	&	0.652	&	0.605	&	0.654	\\
SL	&	0.640	&	0.644	&	\textbf{0.705}	&	0.632	&	0.648	&	0.657	&	0.613	&	\textbf{0.713}	\\
SVM	&	0.652	&	0.669	&	0.668	&	0.668	&	0.652	&	0.674	&	0.588	&	0.675	\\
\hline																	\\
$\mathrm{CNN_{m}}$	&	0.655	&	0.665	&	0.694	&	0.634	&	0.665	&	0.677	&	\textbf{0.704}	&	0.601	\\
$\mathrm{CNN_{ml}}$	&	0.686	&	0.693	&	0.694	&	\textbf{0.680}	&	0.690	&	0.698	&	\textbf{0.704}	&	0.593	\\
$\mathrm{CNN_{mld}}$	&	0.638	&	0.650	&	0.694	&	0.617	&	0.649	&	0.662	&	\textbf{0.704}	&	0.606	\\
$\mathrm{CBLSTM_{m}}$	&	0.694	&	0.682	&	0.686	&	0.635	&	\textbf{0.694}	&	0.692	&	0.562	&	0.582	\\
$\mathrm{CBLSTM_{ml}}$	&	\textbf{0.696}	&	0.697	&	0.686	&	0.653	&	0.689	&	0.684	&	0.562	&	0.581	\\
$\mathrm{CBLSTM_{mld}}$	&	0.645	&	0.643	&	0.686	&	0.602	&	0.655	&	0.656	&	0.562	&	0.588	\\
$\mathrm{LSTM_{m}}$	&	0.686	&	\textbf{0.702}	&	0.685	&	0.653	&	0.688	&	\textbf{0.710}	&	0.557	&	0.589	\\
$\mathrm{LSTM_{ml}}$	&	0.650	&	0.673	&	0.685	&	0.647	&	0.660	&	0.683	&	0.557	&	0.583	\\
$\mathrm{LSTM_{mld}}$	&	0.646	&	0.618	&	0.685	&	0.614	&	0.655	&	0.632	&	0.557	&	0.609	\\
$\mathrm{BLSTMCNN_{m}}$	&	0.684	&	0.683	&	0.678	&	0.568	&	0.682	&	0.689	&	0.685	&	0.576	\\
$\mathrm{BLSTMCNN_{ml}}$	&	0.658	&	0.675	&	0.678	&	0.618	&	0.668	&	0.680	&	0.685	&	0.614	\\
$\mathrm{BLSTMCNN_{mld}}$	&	0.661	&	0.625	&	0.678	&	0.596	&	0.668	&	0.639	&	0.685	&	0.592	\\
\hline
\end{tabular}
\vspace{1mm}
\caption{Cross-domain scores for W00 and WFMALL  datasets as training and test sets, respectively.}\label{cross-domain-tab-w00-wfm}
\end{table}
\subsubsection{Cross-domain results}
\begin{figure}[p]
\centering
\includegraphics[scale=0.9]{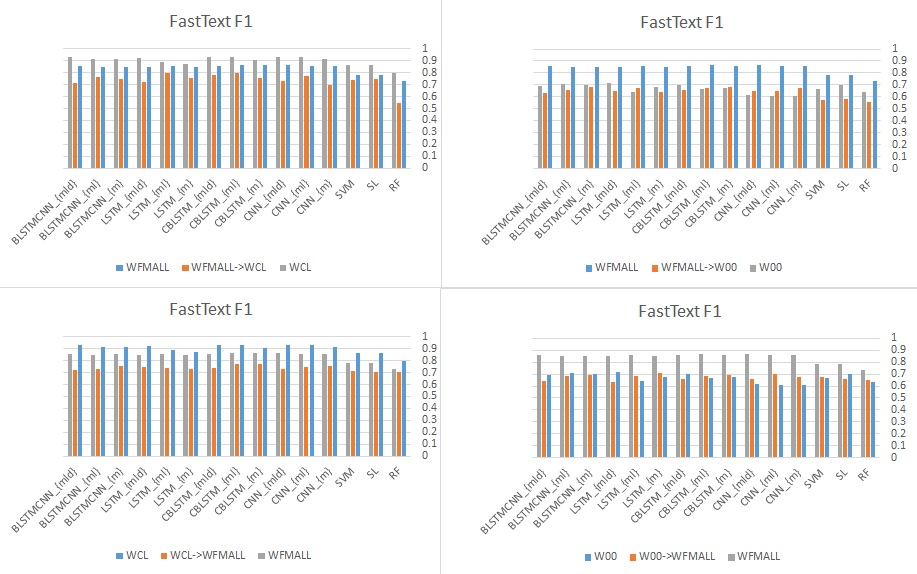}
\vspace{10mm}
\caption{\label{cross-domain}Cross-domain performance. Comparisons between in-domain application and cross-domain application of the models trained on the WFMALL, WCL, and W00 datasets, using fastText embedding and F1 score.}
\end{figure}

The third series of tests was performed using one of the datasets as a training set and another as a test set, where WFMALL is either test or training  dataset.
The primary aim of these tests was to see how well mathematical definitions could be located using general datasets as a training set. An additional goal was to determine whether training the system on mathematical definitions improved recognition of general definitions. 
Results are given in Tables~\ref{cross-domain-tab-wfm-wcl} through  \ref{cross-domain-tab-w00-wfm} and visualized in Figure~\ref{cross-domain}. 
The outcomes that are observed in this experiment are very similar to what was observed during in-domain runs. The deep neural network models that used three representations (see Section~\ref{input-sec}), 
significantly outperformed all the baselines in most cases, with few exceptions. 
From Tables~\ref{cross-domain-tab-wfm-wcl} and \ref{cross-domain-tab-wfm-w00} it can be also seen that models having a CNN layer  
outperform other models in most scenarios. 
As such, we reach the same conclusions as for the in-domain experiments: (1) the CNN model---pure or integrated with BLSTM---achieves better performance than LSTM, which can be explained by the ability of CNN to learn features and reduce the number of free parameters in a high-dimensional  sentence representation; (2) most models usually gain better performance with dependency information.  

As in other evaluation scenarios, sometimes  baselines performed better than NN models, when the sentences were represented with BERT or self-embedding vectors. In general, we observed that BERT and self-embedding vectors did not improve NN performance, in any evaluation scenario. 

Another important observation can be made if we compare between performance of in-domain and cross-domain experiments (see Figure~\ref{cross-domain}). The performance of all models with cross-domain learning is far lower than their performance with in-domain learning. As such, we can conclude that mathematical definitions require special treatment, while using cross-domain learning is inefficient. Models trained to detect general definitions are not sufficient for the detection of mathematical definitions. Likewise, models trained on mathematical domain are not very helpful for detection of general definitions.  

\subsubsection{Binary classification with fine-tuned BERT}
We performed a small experiment with BERT, fine-tuned on our WFMALL dataset and the DE (as a binary classification) task. The purpose of this experiment was to see how the fine-tuning of BERT can improve its performance over the pre-trained one. 

We have used the StrongIO  package~\citep{strongio}  to fine-tune BERT model. We have adapted the code to handle binary sentence classification and train-validation-test mode of evaluation (original code supports train-validation mode only). The BERT model we used is bert$\_$uncased$\_$L-12$\_$H-768$\_$A-12~\citep{bert-code}.\footnote{according to Google release notes~\citep{huggingface} this is the model that can be fine tuned with server having less than 64 GB memory, which fits our server}  This BERT model  has $110,302,011$ parameters, out of which $21,460,737$ are trainable. We have used $75\%$-$5\%$-$25\%$ train-validation-test split as in all other experiments.
\begin{table}
\scriptsize
\begin{tabular}{llllll}
\hline		
Dataset	&	Domain	&	accuracy	&	recall	&	precision	&	F1	\\
\hline
WCL	&	in-domain	&	\textbf{0.964}	&	0.964	&	0.948	&	\textbf{0.956}	\\
WFMALL	&	in-domain	&	0.844	&	0.864	&	0.627	&	0.727	\\
W00	&	in-domain	&	0.804	&	0.854	&	0.489	&	0.622	\\
MULTI	&	multi-domain	&	0.804	&	0.854	&	0.489	&	0.622	\\
WFMALL$=>$WCL	&	cross-domain	&	\textbf{0.929}	&	0.876	&	0.955	&	\textbf{0.914}	\\
WFMALL$=>$W00	&	cross-domain	&	0.666	&	0.000	&	0.000	&	0.000	\\
WCL$=>$WFMALL	&	cross-domain	&	0.516	&	0.391	&	0.960	&	0.555	\\
W00$=>$WFMALL	&	cross-domain	&	0.675	&	0.402	&	0.067	&	0.114	\\
\hline
\end{tabular}
\vspace{1mm}
\caption{Fine-tuned BERT.}\label{bert}
\end{table}

As it can be seen from Table~\ref{bert}, fine-tuned BERT model outperforms other models on WCL dataset in both in-domain and cross-domain scenarios and in both metrics. However, it does not outperform top models on other datasets in any of three scenarios. BERT model has difficulty to detect true positive (definitions) and therefore suffers from low precision, recall, and F1-measure results. 
Therefore, we conclude that even fine-tuned BERT does not perfectly suite the DE task for mathematical domain. 
\subsection{Discussion and error analysis}
As can be observed from the experimental results, the NN models outperform all baselines in most cases of two scenarios (in-domain and multi-domain).

We also can confirm that the use of syntactic information as a part of input configurations ($\mathit{ml}$ and $\mathit{mld}$) improves the results in most scenarios. 
Moreover, during experiments we observed that when dependency encoding is used, keeping separate information about sentence words (as word vectors) has no significant impact on classification results. This allows us to decrease the model size and achieve better time complexity without significantly harming performance.

We can conclude that mathematical definitions require special treatment. 

Finally, we see that generally both CNN and its combination with LSTM are more beneficial than selecting a plain LSTM model. This is probably due to the ability of CNN to perform abstract feature engineering before computing the classification model.

Regarding the word embedding models, we can conclude that neither pre-trained BERT nor self-embedding are helpful for the DE task. BERT usually helps in IR tasks, where the surrounding context of a word is considered in its representation. However, this quality is not helpful for distinguishing between definitions and regular sentences, especially when the representation trained on a general domain and applied on a  specific one, such as mathematical articles. Self-embedding requires a much larger training set than we could provide in our task. Both word2vec and fastText vectors provided comparable results. According to our expectations, 
the fastText vectors usually provided better results due to the larger body of word coverage that fastText gave to all three datasets. 

As regards the runtime performance, training of NN models took from 1 to 3 hours on our server, depending on the model. A significant portion of that time was spent on dependency parsing. Using the  BERT-as-a-service python package~{bert-service} to  obtain BERT vectors on our input text also took considerable time. As for models, those with CNN as their first layer were much faster, due to the feature space reduction.

We tried to understand which sentences represented difficult cases for our models. During annotation process, we found that multiple sentences were assigned different labels by different annotators. Finally, the label for such sentences was decided by majority voting, but all annotators agreed that the decision was not unambiguous. Based on our observation and manual analysis,  we believe that most of the false positive and false negative cases were created by such sentences. We categorized these sentences to the following cases:
\begin{enumerate}
    \item Sentences describing properties of a mathematical object. Example (annotated\footnote{gold standard label = ``definition''} as \textit{definition}
    ):
$$\begin{array}{l}
\mbox{\it An extremum may be local ( a.k.a. a relative extremum; }\\
\mbox{\it an extremum in a given region which is not the overall }\\
\mbox{\it maximum or minimum ) or global. }%
\end{array}$$
We did not instruct our annotators regarding labeling this sentence type and let them make decisions based on their knowledge and intuition. As result, this sentence type received different labels from different annotators. 
\item Sentences providing alternative naming of a known (and previously defined) mathematical object. Example (annotated as \textit{non-definition}
):
$$\begin{array}{l}
\mbox{\it Exponential growth is common in physical processes such }\\
\mbox{\it as population growth in the absence of predators or }\\
\mbox{\it resource restrictions (where a slightly more general form }\\
\mbox{\it is known as the law of growth).}
\end{array}$$
We received the same decisions and the same outcomes in our dataset with this sentence type as with type (1). 
\item Formulations -- sentences that define some mathematical object by a formula (in contrast to a verbal definition, that explains the object's meaning). Example (annotated as \textit{non-definition}
):
$$\begin{array}{l}
\mbox{\it Formulas expressing trigonometric functions of an angle 2x}\\
\mbox{\it  in terms of functions of an angle x, sin(2x) = [FORMULA]. }
\end{array}$$
If both definition and formulation sentences for the same object were provided, our annotators usually assigned them different labels. However, rarely a mathematical object can be only defined by a formula. Also, sometimes it can be defined by both, but the verbal definition is not provided in an analyzed article. In such cases, annotators assigned the \enquote*{definition} label to the formulation sentence. 
\item Sentences that are parts of a multi-sentence definition. Example (annotated as \textit{non-definition}
):
$$\begin{array}{l}
\mbox{\it This polyhedron is known as the dual, or reciprocal. }
\end{array}$$
We instructed our annotators not to assign  \enquote*{definition} label to sentences that do not contain comprehensive information about a defined object. However, some sentences were still annotated as \enquote*{definition,} especially when they appear in a sequence. 
\item Descriptions -- sentences that describe mathematical objects but do not define them unequivocally. Example (annotated as \textit{non-definition}
): 
$$\begin{array}{l}
\mbox{\it A dragon curve is a recursive non-intersecting curve }\\
\mbox{\it  whose name derives from its resemblance to a certain}\\
\mbox{\it  mythical creature. }
\end{array}$$
Although this sentence looks like a legitimate definition (grammatically), it was labeled as non-definition because its claim does not hold in both directions (not every recursive non-intersecting curve is a dragon curve). Because none of our annotators was expert in all mathematical domains, it was difficult for them to assign the correct label in all similar cases. 
\end{enumerate}
As result of subjective  annotation (which occurs frequently in all IR-related areas), none of the ML models trained on our training data were very precise with the ambiguous cases like those described above.  
Below are several examples of sentences misclassified as definitions (false positives\footnote{with gold standard label ``non-definition'' but classified as ``definition''}), from each type described in the list above:
\begin{enumerate}
    \item Property description:
    $$\begin{array}{l}
        \mbox{\it Every pentagonal number is 1/3 of a triangular number.}
     \end{array}$$
    \item Alternative naming:
    $$\begin{array}{l}
        \mbox{\it However, being in \enquote*{one-to-one correspondence} is synonymous }\\
        \mbox{\it  with being a bijection.}
     \end{array}$$
    \item Formulations and notations:
    $$\begin{array}{l}
        \mbox{\it The binomial distribution is therefore given by $P_p(n|N ) = $[FORMULA].}\\
     \:\: \\
     \mbox{\it For a binary relation R, one often writes aRb to mean that $(a, b)$ is in $R \times R$.}
     \end{array}$$
    \item Partial definition:
    $$\begin{array}{l}
        \mbox{\it A center X is the triangle centroid of its own pedal }\\
    \mbox{\it  triangle iff it is the symmedian point.}
    \end{array}$$
     This sentence was annotated as non-definition, because it does not define the symmedian point. 
     \item Description:
     $$\begin{array}{l}
        \mbox{\it The cyclide is a quartic surface, and the lines of curvature }\\
        \mbox{\it on a cyclide are all straight lines or circular arcs.}
    \end{array}$$
\end{enumerate}
Most misclassified definitions (false negatives) can be described by an atypical grammatical structure. Examples of such  sentences can be seen below:
$$\begin{array}{l}
        \mbox{\it Once one countable set S is given, any other set which can be put }\\
        \mbox{\it into a one - to - one correspondence with S is also countable .}\\
        \:\: \\
        \mbox{\it The word cissoid means \enquote*{ivy-shaped .}}\\
        \:\: \\
        
        \mbox{\it A bijective map between two metric spaces that preserves distances, }\\
        \mbox{\it i.e., d(f(x), f(y)) = d(x, y), where f is the map and d(a, b)}\\
        \mbox{\it is the distance function.}
    \end{array}$$
    
We propose to deal with some of the identified error sources as follows. Partial definitions can probably be  discarded by applying part-of-speech tagging and pronouns detection. Coreference resolution (CR) can be used for identification of the referred mathematical entity in a text. Also, the partial definitions problem should be resolved by reduction of the DE task to multi-sentence DE. 
Formulations and notations can probably be discarded by measuring the ratio between mathematical symbolism and regular text in a sentence. Sentences providing alternative naming for mathematical objects can be discarded if we are able to detect the truth definition and then select it from multiple candidates. It can also be probably resolved with the help of such types of CR as split antecedents and coreferring noun phrases. 
\section{Conclusions\label{conc-sec}}
In this paper we introduce a framework for DE from mathematical texts, using deep neural networks. We introduce a new annotated dataset of mathematical definitions, called WFMALL. We test state-of-the-art approaches for the DE task on this dataset. In addition, we introduce a novel representation for text sentences, based  on dependency information, and models with different combinations of CNN and BLSTM layers, and compare them to state-of-the-art results. 

With regard to our hypotheses, we can conclude the following.
\begin{enumerate}
    \item NNs generally perform better than baselines (hypothesis 1 is accepted);
    \item Our experiments demonstrate the superiority of CNN and its combination with LSTM, applied on a syntactically-enriched input representation (hypotheses 2 and 3 are accepted);  
    \item FastText embedding vectors contribute to better NNs performance (hypothesis 4 is accepted);
    \item NNs with self-trained embedding vectors performed the worst, as expected (hypothesis 5 is accepted);
    \item Mathematical definitions require special treatment -- models trained on non-mathematical domains are not very helpful for extraction of mathematical definitions (hypothesis 6 is accepted); 
    \item An additional experiment performed on our newly collected dataset of Wikipedia articles from the mathematical category demonstrates the ability of our approach to detect mathematical definitions in most of the collected articles. However, not every article contains definitions (hypothesis 7 is rejected). 
\end{enumerate}
In addition, we can conclude that using BERT vectors pre-trained on a general context does not gain much performance. Despite our expectations, fine-tuned BERT on the DE classification task and the definitions data  did not demonstrate superiority on mathematical domain. 

In our future work, we plan to extend syntactic structure-based representation from single sentences to the surrounding text. We also plan to expand the classic single-sentence DE task to the task of multi-sentence DE. 
The proposed approach can be adapted to the multi-sentence definition extraction, if we train our model to detect the definition boundaries. This task can be reduced to a multi-class classification, where each sentence must be assigned into one of several classes, such as \enquote*{start of a definition,} \enquote*{definition,} \enquote*{end of definition,} and \enquote*{non definition.} 

\newpage
\bibliographystyle{model5-names}

\bibliography{eswa_mde}

\newpage
\section*{Appendix}
\subsection*{Annotation instructions}
Our annotators were provided with manually selected examples with definitions and non-definitions and asked not to select multi-sentence definitions. We did not provide exact rules describing what is definition and what is not, because if it would be possible, we could easily extract definitions by a rule-based unsupervised classifier.
Below one can see some examples of definitions and non-definitions, provided to the annotators:\\
Definitions:
\begin{itemize}
    \item If two numbers b and c have the property that their difference b - c is integrally divisible by a number m ( i.e., ( b - c)/m is an integer ), then b and c are said to be ``congruent modulo m''.
    \item When the reals are acting, the system is called a continuous dynamical system, and when the integers are acting, the system is called a discrete dynamical system.
    \item In general, an icosidodecahedron is a 32-faced polyhedron.
\end{itemize}
Non-definitions:
\begin{itemize}
    \item It is also called a logical matrix, binary matrix, relation matrix, or Boolean matrix.
    \item It is also known as the hypericosahedron or hexacosichoron.
    \item Each Boolean function has a unique representation (up to order) as a union of complete products.
\end{itemize}
\subsection*{Wikipedia experiment}
We performed an additional experiment that brings an interesting insight into the domain of Wikipedia mathematics-oriented articles. We hypothesized that an efficient approach for definition detection should find at least one definition in every mathematics-oriented document. For testing this hypothesis, we downloaded English Wikipedia dump from \url{https://dumps.wikimedia.org/enwiki/latest/} on Dec 03 2019 and have used this data to extract mathematical articles. This specific snapshot is now available at \url{https://dumps.wikimedia.org/enwiki/20191201/}.

We have parsed this data using a modification of code at \url{https://github.com/thunlp/paragraph2vec/blob/master/gensim/corpora/wikicorpus.py}. Modification included punctuation marks preservation, eliminating supplementary Wiki data and pruning by category name. We only kept categories whose names start with \enquote*{Category: mathematics.} We skipped categories whose names start with \enquote*{Category: mathematics education} because the latter are dedicated to the history of mathematical education and do not contain mathematical definitions. 
Of course, not every article in these categories is a mathematical article containing proper definitions. 
The resulting dataset is denoted by WikiMath.
As a result, we have obtained 3,567 Wikipedia articles containing 110,047 sentences in total.

We tested our hypothesis with one of our best models trained on combined dataset (MULTI) --  $\mathrm{CBLSTM_{ml}}$ with words represented by  fastText word vectors. The model is applied on each sentence from every article. We assign to each sentence  a definition or a non-definition label. 

This is a slow process because dependency parsing is required for all the sentences in WikiMath. We have finished prediction for 1,250 out of 3,567 articles containing 27,318 sentences in total.

According to the prediction results, 760 out of 1,250 articles (around $60\%$) have been found to contain definitions, and 2,210 out of 27,318 (around $8.1\%$) sentences were classified as definitions. These ratios remained quite consistent as the number of processed sentences increased. 

While analyzing the articles where no definitions were detected (despite our expectations), we discovered that indeed not every Wikipedia article contains definitions. For example, article about \enquote*{A Mathematician's Apology}---a 1940 essay by British mathematician G. H. Hardy\footnote{\url{https://en.wikipedia.org/wiki/A_Mathematician\%27s_Apology})}---does not contain any mathematical definition, and no definition was detected by our DE method. 

Examples of sentences classified as definitions can be seen below. 
$$\begin{array}{l}\mbox{\it Algorithmic topology, or computational topology, is a sub-field of}\\
                  \mbox{\it topology with an overlap with areas of computer science, in particular,}\\
                  \mbox{\it computational geometry and computational complexity theory.}\\\\
                  \mbox{\it Harmonices Mundi (Latin: The Harmony of the World, 1619) is a book}\\
                  \mbox{\it by Johannes Kepler.}\\\\
                  \mbox{\it Musica universalis was a traditional philosophical metaphor that}\\
                  \mbox{\it was taught in the quadrivium, and was often called the music of the spheres.}
                  \end{array}$$

\subsection*{Figures}
\begin{figure}[H]
\includegraphics[scale=0.8]{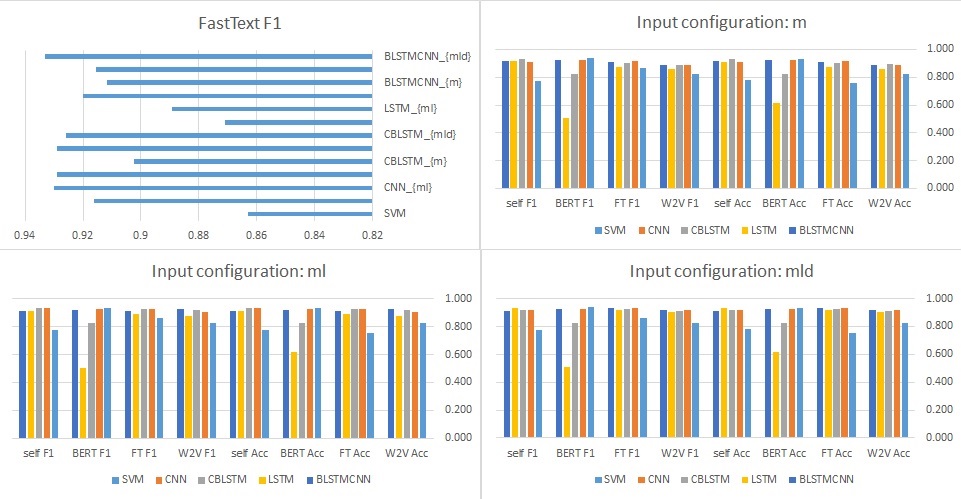}
\vspace{26mm}
\caption{\label{in-domain-wcl}In-domain performance. WCL dataset. Upper left: comparisons between different input configurations across different models with fastText F1. Upper right, bottom left, bottom right: comparisons between different models and the best baseline (SVM) using m, ml, and mld input configurations, respectively. This figure demonstrates F1 and accuracy scores of evaluated models, using three input  configurations, on the WCL dataset. It can be seen that there is a strong correlation between both metrics--F1 and accuracy. Also, it is easy to see that the same distribution of scores holds for all three input configurations. The upper left chart shows that configurations with dependency features (ml and mld) gain better scores. The same holds on other datasets. Therefore, we further report F1 scores only for mld input configuration.}
\end{figure} 
\begin{figure}[p]
\centering
\includegraphics[scale=0.8]{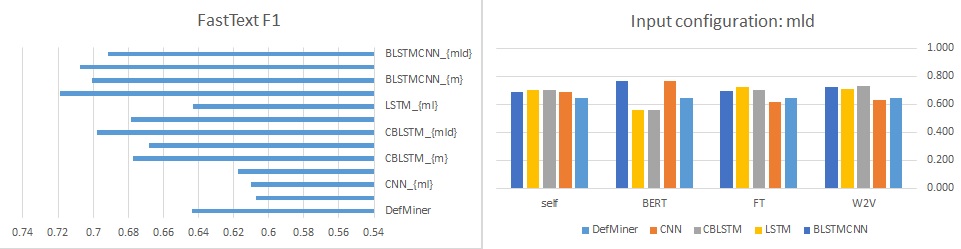}
\vspace{10mm}
\caption{\label{in-domain-w00}In-domain performance. W00 dataset. Left: comparisons between different input configurations across different models with fastText F1. Right: comparisons between different models and the best baseline (DefMiner) using mld input configuration, F1 scores. }
\end{figure} 

\begin{figure}[p]
\centering
\includegraphics[scale=0.8]{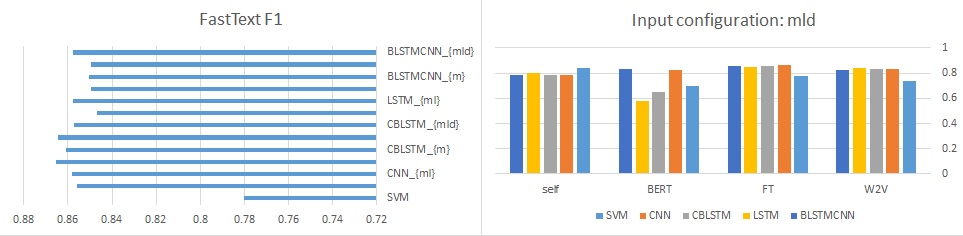}
\vspace{10mm}
\caption{\label{in-domain-wfm}In-domain performance. WFMALL dataset. Left: comparisons between different input configurations across different models with fastText F1. Right: comparisons between different models and the best baseline (SVM) using mld input configuration, F1 scores, respectively.}
\end{figure} 
\begin{figure}[p]
\centering
\includegraphics[scale=0.8]{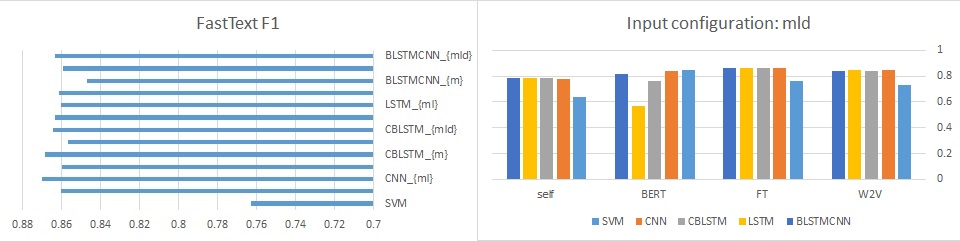}
\vspace{10mm}
\caption{\label{in-domain-multi}In-domain performance. MULTI dataset. Left: comparisons between different input configurations across different models with fastText F1. Right: comparisons between F1 scores of different models and the best baseline (SVM) using mld input configuration. }
\end{figure} 
\end{document}